\documentclass[11pt]{article}
\def\be{\begin{equation}}
\def\ee{\end{equation}}

\newcommand{\ff}[1]{{\mbox{\boldmath $#1$}}}

\def\a{\alpha}

\def\x{\ff{x}}

\begin{document}
\title{Bat Algorithm: Literature Review and Applications}

\author{Xin-She Yang \\
School of Science and Technology, Middlesex University, \\
The Burroughs, London NW4 4BT, United Kingdom.}

\date{{\bf Reference to this article:} Xin-She Yang, Bat algorithm: literature review and applications, {\it Int. J. Bio-Inspired Computation}, Vol.~5, No.~3, pp. 141--149 (2013).
DOI: 10.1504/IJBIC.2013.055093 }

\maketitle

\abstract{Bat algorithm (BA) is a bio-inspired algorithm developed by Yang in 2010
and BA has been found to be very efficient. As a result, the literature has expanded
significantly in the last 3 years. This paper provides a timely review of the bat
algorithm and its new variants. A wide range of diverse applications and case studies are
also reviewed and summarized briefly here. Further research topics are also discussed.
}

\section{Introduction}

Modern optimisation algorithms are often nature-inspired, typically based on
swarm intelligence. The ways for inspiration are diverse and consequently
algorithms can be many different types. However, all these algorithms tend
to use some specific characteristics for formulating the key updating formulae.
For example, genetic algorithms were inspired by Darwinian evolution characteristics
of biological systems, and genetic operators such as crossover and mutation and
selection of the fittest are used. Solutions in genetic algorithms are
represented as chromosomes or binary/real strings.
On the other hand, particle swarm optimisation (PSO) was based on the swarming
behaviour of birds and fish, and this multi-agent system may have emergent
characteristics of swarm or group intelligence (Kennedy and Eberhart, 1995).
Many variants of PSO and improvements exist in the literature, and many new
metaheuristic algorithms have been developed (Cui, 2009;
Yang, 2010; Yang and Deb, 2010; Yang et al., 2011; Yang et al., 2013).

Algorithms such as genetic algorithms and PSO can be very useful, but they
still have some drawbacks in dealing with multimodal optimization problems.
One major improvement is the firefly algorithm (FA) which was based on the
flashing characteristics of tropical fireflies (Yang, 2008). The attraction
behaviour, light intensity coding, and distance dependence provides a surprising
capability to enable firefly algorithm to handle nonlinear, multimodal optimization
problems efficiently. Furthermore, cuckoo search (CS) was based on the brooding behaviour
of some cuckoo species (Yang and Deb, 2009; Gandomi et al, 2013) which was combined with L\'evy flights.
The CS algorithm is efficient because it has very good convergence behaviour
that can be proved using Markovian probability theory. Other methods such as
eagle strategy are also very effective (Yang and Deb, 2010; Gandomi et al, 2012).

As a novel feature, bat algorithm (BA) was based on the echolocation features of
microbats (Yang, 2010), and BA uses a frequency-tuning technique to increase
the diversity of the solutions in the population, while at the same, it uses
the automatic zooming to try to balance exploration and exploitation during
the search process by mimicking the variations of pulse emission rates and loudness
of bats when searching for prey. As a result, it proves to be very efficient
with a typical quick start. Obviously, there is room for improvement.
Therefore, this paper intends to review the latest developments of the bat algorithm.
The paper is organized as follows: Section 2 introduces the basic behaviour
of echolocation and the standard formulation of the bat algorithm.
Section 3 provides a brief description of the variants of BA, and Section 4
highlights the diverse applications of bat algorithm and its variants.
Finally, Section 5 provides some discussions and topics for further research.

\section{The Standard Bat Algorithm}

The standard bat algorithm was based on the echolocation or bio-sonar characteristics
of microbats. Before we outline the details of the algorithm, let us briefly
introduce the echolocation.

\subsection{Echolocation of Microbats}

There are about 1000 different species of bats ( Colin, 2000).
Their sizes can vary widely, ranging from the tiny bumblebee bat
of about 1.5 to 2 grams to the giant bats with wingspan of about 2 m
and may weight up to about 1 kg.  Most bats uses echolocation to a certain degree;
among all the species, microbats use
echolocation extensively, while megabats do not.

Microbats typically use a type of sonar, called, echolocation, to detect prey, avoid
obstacles, and locate their roosting crevices in the dark. They can emit
a very loud sound pulse and listen for the echo that bounces back from
the surrounding objects (Richardson, 2008). Their pulses vary in properties and can be
correlated with their hunting strategies, depending on the species.
Most bats use short, frequency-modulated signals to sweep through about an octave,
and each pulse lasts a few thousandths of a second (up to about 8 to 10 ms) in
the frequency range of 25kHz to 150 kHz. Typically,  microbats can emit
about 10 to 20 such sound bursts every second, and the rate
of pulse emission can be sped up to about 200 pulses per second when homing on their prey.
Since the speed of sound in air is about $v=340$ m/s, the wavelength
$\lambda$ of the ultrasonic sound bursts with a constant frequency $f$
is given by $\lambda=v/f$,  which is in the range of 2mm to 14mm for
the typical frequency range from 25kHz to 150 kHz. Interestingly, these wavelengths
are in the same order of their prey sizes.

Though in reality microbats can also use time delay between their ears
and loudness variations to sense three-dimensional surroundings,
we are mainly interested in some features of the echolocation
so that we can some link them with
the objective function of an optimization problem, which
makes it possible to formulate a smart, bat algorithm.

\subsection{Bat Algorithm}

Based on the above description and characteristics of bat echolocation,
Xin-She Yang (2010) developed the bat algorithm with  the following
three idealised rules:

\begin{itemize}

\item[1.] All bats use echolocation to sense distance, and
they also `know' the difference between food/prey and background barriers
in some magical way;

\item[2.] Bats fly randomly with velocity $\ff{v}_i$
at position $\x_i$ with a frequency $f_{\min}$, varying wavelength $\lambda$
and loudness $A_0$ to search for prey. They can automatically adjust
the wavelength (or frequency) of their
emitted pulses and adjust the rate of pulse emission $r \in [0,1]$,
depending on the proximity of their target;

\item[3.] Although the loudness can vary in many ways, we assume that
the loudness varies from a large (positive) $A_0$ to a minimum
constant value $A_{\min}$.

\end{itemize}

For simplicity, we do not use ray tracing in this algorithm, though it can
form an interesting feature for further extension. In general, ray tracing
can be computational extensive, but it can be a very useful feature
for computational geometry and other applications. Furthermore,
a given frequency is intrinsically linked to a wavelength.
For example, a frequency range of [$20$kHz, $500$kHz] corresponds to
a range of wavelengths from $0.7$mm to $17$mm in the air. Therefore,
we can describe the change either in terms of frequency $f$
or wavelength $\lambda$ to suit different applications, depending
on the ease of implementation and other factors.

\subsection{Bat Motion}

Each bat is associated with a velocity $\ff{v}_i^t$ and a location $\x_i^t$,
at iteration $t$, in a $d$-dimensional search or solution space. Among
all the bats, there exists a current best solution $\x_*$. Therefore,
the above three rules can be translated into the updating equations
for $\x_i^{t}$ and velocities $\ff{v}_i^{t}$:
\be f_i =f_{\min} + (f_{\max}-f_{\min}) \beta, \label{f-equ-150} \ee
\be \ff{v}_i^{t} = \ff{v}_i^{t-1} +  (\x_i^{t-1} - \x_*) f_i , \ee
\be \x_i^{t}=\x_i^{t-1} + \ff{v}_i^t,  \label{f-equ-250} \ee
where $\beta \in [0,1]$ is a random vector drawn from a uniform distribution.

As mentioned earlier, we can either use wavelengths or frequencies for
implementation, we will use $f_{\min}=0$ and $f_{\max}=O(1)$, depending on the
domain size of the problem of interest. Initially, each bat is randomly assigned
a frequency which is drawn uniformly from $[f_{\min}, f_{\max}]$.
For this reason, bat algorithm can be considered as a frequency-tuning algorithm
to provide a balanced combination of exploration and exploitation.
The loudness and pulse emission rates essentially provide a mechanism for
automatic control and auto zooming into the region with promising solutions.

\subsection{Variations of Loudness and Pulse Rates}

In order to provide an effective mechanism to control the exploration and exploitation
and switch to exploitation stage when necessary, we have to vary the loudness $A_i$
and the rate $r_i$ of pulse emission during the iterations.
Since the loudness usually decreases once a bat has found its prey,
while the rate of pulse emission increases, the loudness can be chosen as any value of convenience,
between $A_{\min}$ and $A_{\max}$, assuming $A_{\min}=0$ means that a bat
has just found the prey and temporarily stop emitting any sound.
With these assumptions, we have
\be A_i^{t+1}=\alpha A_{i}^{t}, \;\;\;\;\; r_i^{t+1}= r_i^0 [1-\exp(-\gamma t)],
\label{rate-equ-50} \ee
where $\alpha$ and $\gamma$ are constants. In essence, here $\alpha$ is similar
to the cooling factor of a cooling schedule in simulated annealing.
For any $0<\alpha<1$ and $\gamma>0$, we have
\be A_i^t \rightarrow 0, \;\;\; r_i^t \rightarrow r_i^0, \;\;\textrm{as} \;\;
t \rightarrow \infty. \ee
In the simplest case, we can use $\a=\gamma$,
and we have used $\a=\gamma=0.9$ to $0.98$ in our simulations.

\section{Variants of Bat Algorithm}

The standard bat algorithm has many advantages, and one of the key advantages is
that it can provide very quick convergence at a very initial stage by switching
from exploration to exploitation. This makes it an efficient algorithm
for applications such as classifications and others when a quick solution
is needed. However, if we allow the algorithm to switch to exploitation
stage too quickly by varying $A$ and $r$ too quickly, it may lead to stagnation
after some initial stage. In order to improve the performance, many methods and
strategies have been attempted to increase the diversity of the solution and thus
to enhance the performance, which produced a few good variants of bat algorithm.

From a quick literature survey, we found the following bat algorithm variants:
\begin{itemize}

\item {\it Fuzzy Logic Bat Algorithm (FLBA)}: Khan et al. (2011) presented a variant by introducing fuzzy logic
into the bat algorithm, they called their variant fuzzy bat algorithm.

\item {\it Multiobjective bat algorithm (MOBA)}: Yang (2011) extended BA to deal with multiobjective optimization, which has demonstrated its effectiveness for solving a few design benchmarks
    in engineering.

\item {\it K-Means Bat Algorithm (KMBA)}: Komarasamy and Wahi (2012) presented a combination of K-means and bat algorithm (KMBA) for efficient clustering.

\item {\it Chaotic Bat Algorithm (CBA)}: Lin et al. (2012) presented a chaotic bat algorithm using L\'evy flights and chaotic maps to carry out parameter estimation in dynamic biological systems.

\item {\it Binary bat algorithm (BBA)}: Nakamura et al. (2012) developed a discrete version of bat algorithm to solve classifications and feature selection problems.

\item {\it Differential Operator and L\'evy flights Bat Algorithm (DLBA)}:  Xie et al. (2013) presented a variant of bat algorithm using differential operator and L\'evy flights to solve function optimization problems.

\item {\it Improved bat algorithm (IBA)}: Jamil et al. (2013) extended the bat algorithm with a good combination of L\'evy flights and subtle variations of loudness and pulse emission rates. They tested the IBA versus over 70 different test functions and proved to be very efficient.

\end{itemize}

There are other improvements and variants of bat algorithm. For example, Zhang and Wang (2012) used mutation to enhance the diversity of solutions
and then used for image matching. In addition, Wang and Guo (2013) hybridized bat algorithm with harmony search and have produced a hybrid bat algorithm for numerical optimization of function benchmarks.

On the other hand, Fister Jr et al. (2013) developed a hybrid bat algorithm using differential
evolution as a local search part of bat algorithm, while Fister et al. (2013) incorporate
quaternions into bat algorithm and presented a quaternion bat algorithm (QBA) for computational
geometry and large-scale optimization problems with extensive rotations.
It can be expect that more variants are still under active research.

\section{Applications of Bat Algorithm}

The standard bat algorithm and its many variants mean that the applications are also very diverse. In fact, since the original bat algorithm
has been developed (Yang, 2010), Bat algorithms have been applied in almost every area of optimization, classifications, image processing, feature selection,
scheduling, data mining and others. In the rest of the paper, we will briefly highlight some of the applications (Yang, 2010;Parpinelli and Lopes, 2011;
Yang et al., 2012a; Yang, 2012; Yang, 2013; Gandomi et al., 2013).

\subsection{Continuous Optimization}

Among the first set of applications of bat algorithm, continuous optimization in the context of engineering design optimization has been
extensively studied, which demonstrated that BA can deal with highly nonlinear problem efficiently and can find the optimal
solutions accurately (Yang, 2010; Yang and Gandomi, 2012; Yang, 2012; Yang et al., 2012a). Case studies include pressure vessel design, car side design,
spring and beam design, truss systems, tower and tall building design and others.  Tsai et al. (2011) solved numerical optimization
problems using bat algorithm.

In addition, Bora et al. (2012) optimized the brushless DC wheel motors using bat algorithm with superior results.
BA can also handle multiobjective problems effectively (Yang, 2011).

\subsection{Combinatorial Optimization and Scheduling}

From computational complexity point of view, continuous optimization problems can be considered as easy, though it may be
still very challenging to solve. However, combinatorial problems can be really hard, often non-deterministic
polynomial time hard (NP-hard). Ramesh et al. (2013) presented a detailed study of combined economic
load and emission dispatch problems using bat algorithm. They compared
bat algorithm with ant colony algorithm (ABC), hybrid genetic algorithm and
other methods, and they concluded that bat algorithm is easy to implement and
much superior to the algorithms in terms of accuracy and efficiency.

Musikapun and Pongcharoen (2012) solved multi-stage, multi-machine, multi-product
scheduling problems using bat algorithm, and they solved a class of non-deterministic
polynomial time (NP) hard problems with a detailed parametric study. They also implied that
that the performance can be further improved by about 8.4\% using optimal set of parameters.

\subsection{Inverse Problems and Parameter Estimation}

Yang et al. (2012b) use the bat algorithm to study topological shape optimization
in microelectronic applications so that materials of different thermal properties can be
placed in such a way that the heat transfer is most efficient under stringent constraints.
It can also be applied to carry out parameter estimation as an inverse problem. If an inverse
problem can be properly formulated, then bat algorithm can provide better results than
least-squares methods and regularization methods.

Lin et al. (2012) presented a chaotic L\'evy flight bat algorithm to estimate
parameters in nonlinear dynamic biological systems, which proved the effectiveness
of the proposed algorithm.

\subsection{Classifications, Clustering and Data Mining}

Komarasamy and Wahi (2012) studied K-means clustering using bat algorithm
and they concluded that the combination of both K-means and BA can achieve
higher efficiency and thus performs better than other algorithms.

Khan et al. (2011) presented a study of a clustering problem for office workplaces using
a fuzzy bat algorithm. Khan and Sahari (2012a)
also presented a comparison study of bat algorithm with
PSO, GA, and other algorithms in the context for e-learning, and thus suggested
that bat algorithm has clearly some advantages over other algorithms. Then,
they (Khan and Sahari, 2012b) also presented a study of clustering problems using bat
algorithm and its extension as a bi-sonar optimization variant with good results.

On the other hand, Mishra et al. (2012) used bat algorithm to classify microarray data,
while Natarajan et al. (2012) presented a comparison study of cuckoo search and bat
algorithm for Bloom filter optimization. Damodaram and Valarmathi (2012) studied
phishing website detection using modified bat algorithm and achieved very good results.

Marichelvam and Prabaharan (2012) used bat algorithm to study hybrid flow shop scheduling
problems so as to minimize the makespan and mean flow time. Their results suggested
that BA is an efficient approach for solving hybrid flow shop scheduling problems.
Faritha Banu and Chandrasekar (2013) used a modified bat algorithm to record
deduplication as an optimization approach and data compression technique. Their
study suggest that the modified bat algorithm can perform better than genetic programming.

\subsection{Image Processing}

Abdel-Rahman et al. (2012) presented a study for full body human pose estimation
using bat algorithm, and they concluded that BA performs better than
particle swarm optimization (PSO), particle filter (PF) and annealed particle filter (APF).

Du and Liu (2012) presented a variant of bat algorithm with mutation for image matching,
and they indicated that their bat-based model is more effective and feasible in imagine matching
than other models such as differential evolution and genetic algorithms.

\subsection{Fuzzy Logic and Other Applications}

Reddy and Manoj (2012) presented a study of optimal capacitor placement
for loss reduction in distribution systems using bat algorithm. It
combines with fuzzy logic to find optimal capacitor sizes so as to
minimize the losses. Their results suggested that the real power loss
can be reduced significantly.

Furthermore, Lemma et al. (2011) used fuzzy systems and bat algorithm for exergy
modelling, and later Tamiru and Hashim (2013) applied bat algorithm to study fuzzy
systems and to model exergy changes in a gas turbine.

At the time of writing when we searched the Google scholar and other databases, we found other papers on bat algorithm
that were either just accepted or conference presentations. However, there is not enough
detail to be included in this review. In fact, as the literature is expanding,
more and more papers on bat algorithm are emerging, a further timely review will be
needed within the next two years.

\section{Discussions and Conclusions}

Likely many metaheuristic algorithms, bat algorithm has the advantage of simplicity
and flexibility. BA is easy to implement, and such a simple algorithm can be
very flexible to solve a wide range of problems as we have seen in the above review.

\subsection{Why Bat Algorithm is Efficient}

A natural question is: why bat algorithm is so efficient? There are many reasons for
the success of bat-based algorithms. By analysing the key features and updating equations,
we can summarize the following three key points/features:

\begin{itemize}
\item Frequency tuning: BA uses echolocation and frequency tuning to solve problems.
Though echolocation is not directly used to mimic the true function in reality,
frequency variations are used. This capability can provide some functionality that may be
similar to the key feature used in particle swarm optimization and harmony search.
Therefore, BA possess the advantages of other swarm-intelligence-based algorithms.

\item Automatic zooming: BA has a distinct advantage over other metaheuristic algorithms. That is,
BA has a capability of automatically zooming into a region where promising solutions have been found.
This zooming is accompanied by the automatic switch from explorative moves to local
intensive exploitation. As a result, BA has a quick convergence rate, at least at early stages of
the iterations, compared with other algorithms.

\item Parameter control: Many metaheuristic algorithms used fixed parameters by using some, pre-tuned
algorithm-dependent parameters. In contrast, BA uses parameter control, which can vary the values
of parameters ($A$ and $r$) as the iterations proceed. This provides a way to automatically
switch from exploration to exploitation when the optimal solution is approaching. This gives
another advantages of BA over other metaheuristic algorithms.

\end{itemize}
In addition, preliminary theoretical analysis by Huang et al.(2013) suggested that BA has
guaranteed global convergence properties under the right condition, and BA can also solve large-scale
problems effectively.

\subsection{Further Research Topics}

However, there are still some important issues that require more research. These key issues are:
parameter-tuning, parameter control and speedup of convergence.

Firstly, parameter-tuning is important for any metaheuristic algorithm to work properly.
In almost all cases, the performance of an algorithm is largely dependent on the parameters
of the algorithm. To find the best parameter settings, detailed parametric studies have to
be carried out. It is not known yet if there is a method to automatically tune parameters
for an algorithm to achieve the optimal performance for a given set of problems. This should
be an important topic for further research.

Secondly, associated with the parameter tuning, there is an important issue of parameter control.
In many algorithms, the parameter settings are fixed, and these settings will not vary during the
iterations. It could be advantageous and sometime necessary to vary the values of algorithm-dependent
parameters during the iterative search process. How to vary or control these parameters
is another, higher level, optimization problem, which needs further studies. For bat algorithm,
we have introduced the basic parameter control strategy, there is still room for improvement.
An open question is that: what is the best control strategy so as to switch from exploration
to exploitation at the right time?

Finally, even though the bat algorithm and other algorithms are efficient, it is still possible
to improve and enhance their performance further. However, how to speed up the convergence
of an algorithm is still a very challenging question. It is hoped this this paper can inspire
more research in the near future. Future research should focus on the theoretical understanding
of metaheuristic algorithms and large-scale problems in real-world applications.

\end{document}